\documentclass[runningheads]{llncs}
\usepackage{graphicx}
\usepackage{amsmath,amssymb} 

\usepackage{color}
\usepackage{comment}
\usepackage{algorithmic}
\usepackage{algorithm}
\usepackage{subfigure}
\usepackage{multirow}

\newcommand{\ttmaxiter}{\texttt{max\_iter}}

\newcommand{\ttnclusters}{\texttt{n\_clusters}}

\newcommand{\cM}{\mathcal{M}}

\newcommand{\cS}{\mathcal{S}}
\newcommand{\cD}{\mathcal{D}}

\newcommand{\cX}{\mathcal{X}}
\newcommand{\cC}{\mathcal{C}}

\newcommand{\bX}{\mathbf{X}}
\newcommand{\bY}{\mathbf{Y}}

\newcommand{\bI}{\mathbf{I}}

\newcommand{\bS}{\mathbf{S}}

\newcommand{\bV}{\mathbf{V}}

\newcommand{\bx}{\mathbf{x}}
\newcommand{\bc}{\mathbf{c}}
\newcommand{\bC}{\mathbf{C}}

\newcommand{\bp}{\mathbf{p}}
\newcommand{\bq}{\mathbf{q}}

\newcommand{\bbR}{\mathbb{R}}

\newcommand{\bOmega}{\mathbf{\Omega}}
\newcommand{\bgamma}{\boldsymbol{\gamma}}
\newcommand{\brho}{\boldsymbol{\rho}}

\begin{document}
\title{A Binary Optimization Approach for Constrained K-Means Clustering\thanks{“This work was supported by an Asian Office of Aerospace Research and Development Grant FA2386-16-1-4027 and an ARC Future Fellowship FT140101229 to MM. Eriksson was supported by FT170100072.”
		}} 
\titlerunning{A Binary Optimization Approach for Constrained K-Means Clustering} 


\author{Huu M. Le\inst{1}\and
Anders Eriksson\inst{1}\and
Thanh-Toan Do\inst{2}\and
Michael Milford\inst{1}}
%

\authorrunning{Huu M. Le et al.} 


\institute{Queensland University of Technology, Australia \\
\email{\{huu.le, anders.eriksson, michael.milford\}@qut.edu.au}\\
\and
University of Liverpool, England \\
\email{thanh-toan.do@liverpool.ac.uk}}

\maketitle

\begin{abstract}
K-Means clustering still plays an important role in many computer vision problems. While the conventional Lloyd method, which alternates between centroid update and cluster assignment, is primarily used in practice, it may converge to solutions with empty clusters. Furthermore, some applications may require the clusters to satisfy a specific set of constraints, e.g., cluster sizes, must-link/cannot-link. 
Several methods have been introduced to solve constrained K-Means clustering. Due to the non-convex nature of K-Means, however, existing approaches may result in sub-optimal solutions that poorly approximate the true clusters. In this work, we provide a new perspective to tackle this problem by considering constrained K-Means as a special instance of Binary Optimization. We then propose a novel optimization scheme to search for feasible solutions in the binary domain. This approach allows us to solve constrained K-Means clustering in such a way that multiple types of constraints can be simultaneously enforced. Experimental results on synthetic and real datasets show that our method provides better clustering accuracy with faster run time compared to several existing techniques.
\end{abstract}


\section{Introduction}
Since the early days of computer vision and machine learning research, K-Means~\cite{macqueen1967some} has been shown to be an outstanding algorithm for a wide range of applications involving data clustering, such as image retrieval, segmentation, object recognition, etc. Despite a long history of research and developments, K-Means is still being employed as an underlying sub-problem for many state-of-the-art algorithms. Take vector quantization (VQ)~\cite{gersho2012vector}, a well-known algorithm for approximate nearest neighbor (ANN) search, as an example. In order to learn a code-book containing $k$ code-words from the given dataset, VQ uses K-Means to partition the training data into $k$ non-overlapping clusters (the value of $k$ depends on the applications and storage requirements). The $k$ centroids provided by K-Means are then used as $k$ code-words. During the training process, each data point is then encoded by associating it with the nearest centroid and is stored by the index of the corresponding codeword using $\log_2k$ bits. The success of VQ in nearest neighbor search and image retrieval has inspired many of its variants~\cite{jegou2011product,ge2013optimized,kalantidis2014locally,le2018deepvq}. Among such methods, the task of data clustering is still primarily handled by K-Means, thus the effectiveness of K-Means clustering contributes substantially to the overall performance of quantization techniques.

Although being considered as a prime method underpinning many applications, K-Means is in fact NP-Hard, even for instances in the plane~\cite{mahajan2012planar}. Therefore, the task of searching for its globally optimal solution is almost computationally intractable for datasets containing large number of clusters. As a result, one can only expect to obtain sub-optimal solutions to this problem, given that good initializations of the clusters' centroids are available. Lloyd's algorithm~\cite{macqueen1967some} is the commonly used algorithm for K-Means. Starting from the initial centroids and cluster assignments, which can be obtained from random guesses or by employing different initialization strategies~\cite{pena1999empirical}\cite{khan2004cluster}, it solves K-Means by alternating between updating centroids and assigning data points into updated clusters. This iterative process is repeated until convergence (no more improvements can be made to the clusters). Lloyd's approach has the advantage of being simple and can be easily implemented.

One of the main drawbacks of Lloyd's algorithm, however, lies in its inability to enforce constraints, which may become a disadvantage in certain applications. For example, several photo query systems~\cite{althoff2011balanced} require the number of data points distributed into the clusters to be approximately equal. It's also worth mentioning that due to the dependence on initializations of the conventional Lloyd's approach, this algorithm poses a risk of converging to solutions in which one or many clusters contain zero or very few data points (which are more likely to be outliers). In practice, it has also been empirically shown that by enforcing the constraints on the cluster sizes, one gets a better clustering performance~\cite{bradley2000constrained}~\cite{zhu2010data}. Furthermore, in order to take advantage of a-priori contextual information, it can be beneficial for some applications to enforce must-link/cannot-link constraints (i.e., some particular points must/must not be in the same cluster).
Recently, there have been much interests in developing deep learning approaches that require the use of K-Means~\cite{fard2018deep,yang2016towards}. Such deep networks can also be benefited from a K-Means algorithms with some specific requirements on the clusters.  
Henceforth, a K-Means clustering algorithm that strictly enforces constraints during its iterations is highly desirable for many practical applications.

There have been several works that address K-Means clustering with constraints~\cite{bradley2000constrained,liu2017balanced,zhu2010data,malinen2014balanced,li2009constrained}. A large number of works focus on the special case of balanced clusters~\cite{althoff2011balanced,liu2017balanced,malinen2014balanced}. Among them, several methods propose different heuristic schemes to distribute data points into clusters such that the constraints are satisfied~\cite{wagstaff2001constrained,zhu2010data}. On the other hand,~\cite{bradley2000constrained} solves the assignment step by optimizing the linear assignment problem. The authors in~\cite{malinen2014balanced} proposed to use max-flow to address constrained K-Means. However, their algorithm can only work for problem instances containing a small number of clusters, as it is rather computationally expensive to optimize the min-cut algorithm for a large number of nodes. Balanced K-Means has recently been considered in~\cite{liu2017balanced} where the balance constraint is enforced by solved by minimizing a least squares problem. This approximation scheme requires an annealing parameter, which needs to be separately tuned for each problem instance. Also,~\cite{liu2017balanced} can only handle balance constraints, which makese it inflexible for problems that require different cardinality constraints for different clusters, not to mention the fact that must-link and cannot-link constraints are unable to be enforced in~\cite{liu2017balanced}.

\subsubsection{Contributions} In this paper, we revisit the constrained K-Means clustering problem and propose a novel approach to address it in such a way that different types of constraints can be simultaneously enforced. Particularly, we look at the problem at the perspective of a binary optimization problem and devise an algorithm to optimize it so that all the constraints are strictly enforced at each iteration. Unlike other relaxation schemes, our algorithm aims to find feasible solutions in the binary domain, i.e., the cluster assignments are represented by a binary matrix that satisfies the given constraints. One notable feature of our technique is that different types of constraints can be easily integrated into the proposed framework. The experiment results on multiple synthetic and real datasets show that our technique provides much better clustering quality while the run time is comparable to (or even faster than) other approaches.

\section{Related work}
Among other works on constrained clustering~\cite{bradley2000constrained,liu2017balanced,zhu2010data,malinen2014balanced,li2009constrained}, our method is closely related to the methods proposed in~\cite{bradley2000constrained} and~\cite{wagstaff2001constrained}. In particular,~\cite{bradley2000constrained} concerns the problem of K-Means clustering in which the number of points in cluster $l$ must be greater than $\tau_l$, $l=1,\dots, k$, with the main goal is to prevent K-Means from providing clusters containing zero or very few points. Similar to the conventional K-Means, the algorithm proposed in~\cite{bradley2000constrained} comprises two steps: centroid update and cluster assignment. To address the constraints on the cluster sizes, the cluster assignment step in~\cite{bradley2000constrained} is formulated as a linear programming (LP) problem, in which the requirements for the cluster sizes are expressed as linear constraints on the assignment matrix. In fact, the task of solving for the assignment matrix is a Mixed Integer Program (MIP). However, with the special case in~\cite{bradley2000constrained}, it can be proven that the solutions of LP are guaranteed to be binary, i.e., each element of the assignment matrix returned from LP is guaranteed to be either $0$ or $1$. Therefore, the task of solving a MIP boils down to solving an LP as the integrality of the solutions can always be assured. Due to the relationship between the task of distributing points into the clusters and the linear assignment problem, the constrained clustering with cluster size constraints can also be consider as finding a minimum flow on a network. A balanced K-Means algorithm based on this approach is proposed in~\cite{malinen2014balanced}.

Another type of constraint that finds its use in many clustering problems is the must-link (or cannot-link) constraints~\cite{wagstaff2001constrained}. These constraints require a subset of data points must (or must not) be assigned to the same clusters. They are usually incorporated into the clustering process when prior application domain information is available, with the aim to boost the overall accuracy. Note that the must-link and cannot-link constraints can also be formulated as linear constraints (as will be explained in the later sections). However, the solutions of LP is no longer guaranteed to be binary. A heuristic scheme is proposed by~\cite{wagstaff2001constrained} to tackle the constrained clustering problem, where the points are distributed into the cluster in such a way that the constraint violations are minimized.

Besides~\cite{bradley2000constrained} and~\cite{wagstaff2001constrained}, there are also many other works that address clustering with constraints, with a majority of them focusing on the cluster-size constraints. Our work can be considered as a generalized version of~\cite{bradley2000constrained}, in which the must-link and cannot-link constraints discussed in~\cite{wagstaff2001constrained} can also be integrated into a binary optimization framework. Due to the introduction of additional constraints, the solutions of the LP proposed in~\cite{bradley2000constrained} are no longer guaranteed to be binary. Our method addresses this by proposing an alternating optimization scheme which aims to find the feasible solutions in the binary domain.

\section{Problem formulation}
\subsection{K-Means clustering}
\label{sec:k-means}
Given a set $\cX$ containing $N$ data points $\cX = \{ \bx_j \}_{j=1}^N$, where $\bx_j \in \bbR^d$, the goal of K-Means is to partition $\cX$ into $k$ non-overlapping clusters. Specifically, K-Means finds a set $\cC$ containing $k$ centroids $\cC = \{\bc_i\}_{i=1}^k$,  and an assignment that distributes the points into $k$ distinct groups $\cS_1, \dots, \cS_k$, where $\cS_1 \cup \dots \cup \cS_k = \cX$ and $\cS_i \cap \cS_j = \emptyset\; \forall i\neq j$, that minimizes the \emph{within cluster sum of squares} (WCSS) distortion. Mathematically speaking, K-Means clustering can be formulated as the following optimization problem
\begin{equation}
\label{eq:k_means}
\min_{\cC, \cS_1,\dots,\cS_k} \sum_{i=1}^k \sum_{\bx_j \in \cS_i} ||\bx_j - \bc_i||^2_2.
\end{equation}
\subsection{Constrained K-Means}
In several applications, it is sometimes desirable to impose some particular restrictions on the clustering. For the ease of representation, we focus on the cluster size constraints~\cite{bradley2000constrained,liu2017balanced}, and must-link/cannot-link constraints~\cite{wagstaff2001constrained}, while other types of constraints (assuming that they can be converted into linear constraints), depending on the applications, can also be integrated into our proposed framework in a similar manner. Assume that we are given as a priori $m$ subsets $\cM_1, \dots, \cM_m$, where $\cM_i \subset \cX$ and points belonging to a particular set $\cM_i$ \emph{must} be grouped into the same cluster. Likewise, assume that $h$ subsets $\cD_1,\dots, \cD_h$ are given, where $\cD \subset \cX$ and each set $\cD_i$ contains points that \emph{must not} be in the same cluster.   Let $f$ represent the cluster assignment operation, i.e., $f(\bx)$ is the cluster that $\bx$ is assigned to. Similar to~\eqref{eq:k_means}, constrained K-Means also seeks to find the set of $k$ cluster centroids $\cC$ and the subsets $\cS_i$ that minimize the WCSS distortion with additional constraints on the clusters, which can be mathematically expressed as 
\begin{subequations}
    \label{eq:constrained_k_means}
    \begin{align}
    & \min_{\cC, f, \cS_1,\dots,\cS_k}  && \sum_{i=1}^k \sum_{\bx_j \in \cS_i} ||\bx_j - \bc_i||^2_2, \label{eq:constrained_k_means:obj} \\
    & \text{subject to} && l_i \le |\cS_i| \le u_i, \;\; \forall i = 1\dots k, \label{eq:constrained_k_means:size} \\
    & &&                   f(\bp^t) = f(\bq^t) \;\; \forall (\bp^t, \bq^t) \in \cM_t, \; t=1\dots m,  \label{eq:constrained_k_means:must_link} \\
    & &&                   f(\bp^t)\neq f(\bq^t) \;\; \forall (\bp^t, \bq^t) \in \cD_t, \; \bp^t \neq \bq^t \; \; t=1\dots h,  \label{eq:constrained_k_means:must_not_link}
    \end{align}
\end{subequations}
where the notation $|\;.\;|$ denotes the cardinality of a set, and the pairs $\{(l_i, u_i)\}_{i=1}^k$ represent the lower and upper bound on the sizes of the corresponding clusters. The constraints~\eqref{eq:constrained_k_means:size} limit the sizes of the clusters, while the must-link and cannot-link constraints are reflected by~\eqref{eq:constrained_k_means:must_link} and~\eqref{eq:constrained_k_means:must_not_link}, i.e., $f$ must assign all points in $\cM_t$ to the same cluster and, similarly, points reside in the same $\cD_t$ must be distributed into different clusters.


\section{Constrained K-Means as Binary Optimization}
Before introducing our approach to tackle constrained K-Means, let us first re-formulate~\eqref{eq:constrained_k_means} as a binary optimization problem.  Let $\bX \in \bbR^{d\times N}$ be the matrix containing the set of $N$ data points, where each point $\bx_i \in \bbR^d$ corresponds to the $i$-th column of $\bX$), and $\bC \in \bbR^{d\times k}$ be the matrix that stores the set of $k$ centroids (each centroid lies in one column of $\bC$). Additionally, let the cluster assignments be represented by a matrix $\bS \in \bbR^{k \times N}$, where $\bS_{ij}$ has the value of $1$ if the data point $\bx_j$ is assigned to the $i$-th cluster and $0$ otherwise. The problem~\eqref{eq:constrained_k_means} can now be written as follows
\begin{subequations}
\label{eq:constrained_k_means_01}
\begin{align}
& \min_{\bC, \bS}  && \| \bX  - \bC\bS\|^2_F, \label{eq:constrained_k_means_01:obj}\\
& \text{subject to} && \bS_{ij} \in \{0, 1\} \;\; \forall i,j, \label{eq:constrained_k_means_01:binary} \\
& && \sum_{i=1}^k \bS_{ij} = 1 \;\; \forall j = 1\dots N, \label{eq:constrained_k_means_01:cluster_assign} \\
& && l_i \le \sum_{j=1}^N \bS_{ij} \le u_i \;\; \forall i=1 \dots k, \label{eq:constrained_k_means_01:size_constraint} \\
& && \bS_{ip} = \bS_{iq}, \;\; i = 1\dots k \;\; \forall (\bx_p, \bx_q) \in \cM_t, \; t=1\dots m, \label{eq:constrained_k_means_01:must_link} \\
& && \bS_{ip} + \bS_{iq} \le 1, \;\; i = 1\dots k \;\; \forall (\bx_p, \bx_q) \in \cD_t, \; t=1\dots m. \label{eq:constrained_k_means_01:must_not_link}
\end{align}
\end{subequations}
With the re-arrangements of variables into the matrices $\bX$, $\bC$ and $\bS$, the objective function~\eqref{eq:constrained_k_means:obj} translates to~\eqref{eq:constrained_k_means_01:obj} where $\|.\|_F$ denotes the Frobenius norm of a matrix. The constraint~\eqref{eq:constrained_k_means_01:binary} restricts $\bS$ to be a binary assignment matrix, while the second constraint~\eqref{eq:constrained_k_means_01:cluster_assign} allows each point to be assigned to only one cluster. The set cardinality requirements in~\eqref{eq:constrained_k_means:size} are reflected by the third constraint~\eqref{eq:constrained_k_means_01:size_constraint}, as the sum of all elements on the $i$-th row of $\bS$ is the number of points being assigned to the $i$-th cluster. Finally,~\eqref{eq:constrained_k_means_01:must_link} and~\eqref{eq:constrained_k_means_01:must_not_link} enforce the must-link and cannot-link constraints. The intuition behind~\eqref{eq:constrained_k_means_01:must_link} is that if two data points $\bx_p$ and $\bx_q$ have the same cluster assignment, the $p$-th column and $q$-th column of $\bS$ must be identical. Consequently, if two data points $\bx_p$ and $\bx_q$ belong to the same cluster, the element-wise sum of the $p$-th column and $q$-th column of $\bS$ must contain a value of $2$. Therefore, by restricting the element-wise sum to be less than $2$ as~\eqref{eq:constrained_k_means_01:must_not_link}, the cannot-link constraint can be enforced.

Henceforth, for the sake of brevity, we define $\mathbf{\Omega}$ as the convex domain in which the linear constraints~\eqref{eq:constrained_k_means_01:cluster_assign},~\eqref{eq:constrained_k_means_01:size_constraint},~\eqref{eq:constrained_k_means_01:must_link},~\eqref{eq:constrained_k_means_01:must_not_link} are satisfied, i.e.,
\begin{equation}
    \label{eq:omega_definition}  
    \bOmega = \{ \bS \in \bbR^{k\times N} | ~\eqref{eq:constrained_k_means_01:cluster_assign},~\eqref{eq:constrained_k_means_01:size_constraint},~\eqref{eq:constrained_k_means_01:must_link},~\eqref{eq:constrained_k_means_01:must_not_link} \}.  
\end{equation}
With the re-formulation~\eqref{eq:constrained_k_means_01} of the constrained K-Means problem, besides the introduced constraints, other restrictions on point assignments can be explicitly enforced by introducing additional constraints for the matrix $\bS$. Therefore, our proposed method for addressing constrained K-Means by solving~\eqref{eq:constrained_k_means_01} can be generalized to incorporate different types of constraints. 

\section{Optimization Strategy}
In this section, we introduce our optimization technique to tackle the constrained K-Means problem with the binary formulation~\eqref{eq:constrained_k_means_01}. Clearly,~\eqref{eq:constrained_k_means_01} is a non-convex quadratic programming problem with binary constraints, thus solving it optimally is a computationally challenging task. This section introduces a novel alternating optimization approach to solve~\eqref{eq:constrained_k_means_01}. In contrast to other relaxation schemes, we introduce an optimization technique aiming at finding feasible solutions for~\eqref{eq:constrained_k_means_01}, i.e., $\bS \in \{0,1\}^{k\times N}$ and $\bS \in \bOmega$. This allows us to devise a better approximation approach compared to other heuristic or relaxation methods, which will be empirically demonstrated in our experiments.

First, observe that the optimization problem~\eqref{eq:constrained_k_means_01} consists of two sets of variables, namely, $\bC$ and $\bS$, where the constraints are only enforced on the matrix $\bS$. This allows us, similar to the conventional K-Means approach, to set up an optimization strategy that involves alternatively updating $\bC$ and $\bS$ until convergence. Particularly, the update step for $\bC$ with a fixed matrix $\bS$ can be written as
\begin{equation}
\label{eq:C_update}
\begin{aligned}
& \min_{\bC}  && \| \bX  - \bC\bS\|^2_F, \\
\end{aligned}
\end{equation}
while with $\bC$ fixed, updating $\bS$ can be written as
\begin{equation}
\label{eq:S_udpate}
\begin{aligned}
& \min_{\bS}  && \| \bX  - \bC\bS\|^2_F, \\
& \text{subject to} && \bS \in \{0,1\}^{k\times N}, \\
& &&  \bS \in \bOmega.
\end{aligned}
\end{equation}
In the following, we provide more details on the updating steps.

\subsection{Updating the centroids}
To solve~\eqref{eq:C_update}, notice that it is a convex quadratic problem in which solutions can be derived in closed-form. However, to avoid numerical issues of the solutions for large-scale problems, we update $\bC$ by solving the regularized least square problem~\cite{rifkin2007notes}: 
\begin{equation}
\label{eq:C_update_rls}
\begin{aligned}
& \min_{\bC}  && \| \bX  - \bC\bS\|^2_F  \;\; +  \;\; \lambda \|\bC\|_F^2, \\
\end{aligned}
\end{equation}
where $\lambda$ is the regularize parameter. The problem~\eqref{eq:C_update_rls} also admits a closed-form solution
\begin{equation}
\label{eq:C_update_solution}
\begin{aligned}
\bC = \bX \bS^T (\bS\bS^T + \lambda \bI)^{-1}. 
\end{aligned}
\end{equation}
The advantage of~\eqref{eq:C_update_solution} is that $(\bS\bS^T + \lambda \bI)$ is guaranteed to be full-rank, allowing the inversion to be computed efficiently by several matrix factorization techniques. We choose $\lambda$ to be $10^{-4}$ in all the experiments to prevent $\lambda$ from affecting the solution of $\bC$. 

\subsection{Updating the assignment matrix} 
Due to the special structure of $\bS$, the problem of finding the cluster assignments~\eqref{eq:S_udpate} can be written as an optimization problem with linear objective function~\cite{bradley2000constrained}. Let $\bY \in \bbR^{k \times N}$ be a matrix where each element $\bY_{i,j}$ is defined as $\bY_{i,j} = \|\bc_i - \bx_j\|_2^2$ ($\bx_j$ is the $j$-th column of $\bX$ and $\bc_i$ is the $i$-th column of $\bC$). The problem~\eqref{eq:S_udpate} is equivalent to
\begin{subequations}
    \label{eq:S_udpate_linear}
    \begin{align}
    & \min_{\bS}  && \langle \bY, \bS \rangle, \\
    & \text{subject to} && \bS \in \bOmega,\;\; \bS \in \{0,1\}^{k\times N} \label{eq:S_udpate_linear:binary} 
    \end{align}
\end{subequations}
If the binary constraints in~\eqref{eq:S_udpate_linear:binary} are ignored,~\eqref{eq:S_udpate_linear} becomes a convex linear programming (LP) problem. Based on that observation, a straightforward relaxation technique to tackle~\eqref{eq:S_udpate_linear} is to allow $\bS_{ij}$ to be in the continuous domain, i.e., $0 \le \bS_{i,j} \le 1 \;\; \forall i,j$. Then, $\bS$ can be updated efficiently by solving the relaxed version~\eqref{eq:S_udpate_linear}, which can be done by taking advantage of several state-of-the-art LP solvers. However, the solutions returned by LP may not be feasible w.r.t.~\eqref{eq:S_udpate_linear}, i.e., $\exists (i,j), \bS_{i,j} \notin \{0,1\}$. In order to project the LP solutions back to the binary domain of~\eqref{eq:S_udpate_linear}, a thresholding step needs to be executed, which may lead to loose approximation if the threshold is not chosen properly. 

To overcome such drawbacks, we introduce an alternating optimization approach to find the feasible solutions for~\eqref{eq:S_udpate_linear} in an optimal way. Our approach is inspired by several algorithms on feasibility pump~\cite{fischetti2005feasibility,bertacco2007feasibility,geissler2017penalty}, which aim to find feasible solutions for mixed integer programming (MIP) problems. Our formulation allows the problem to be tackled using alternating optimization. As will be shown in the experiments, our algorithm provides a better solution for the constrained K-Means problem compared to other approximation techniques.

Let us first introduce the auxiliary variable $\bV \in \bbR^{k \times N}$ that has the same size as $\bS$. Furthermore, with the constraint that $\bV \in \{0,1\}^{k\times N}$, the problem~\eqref{eq:S_udpate_linear} can now be written equivalently as
\begin{equation}
    \label{eq:S_udpate_SV}
    \begin{aligned}
    & \min_{\bS, \bV}  && \langle \bY, \bS \rangle, \\
    & \text{subject to} && 0 \le \bS_{i,j} \le 1 \;\; \forall i,j, \;\; \bS \in \bOmega, \\
    & && \bV \in \{0,1\}^{k\times N},\;\; \bS = \bV.  
    \end{aligned}
\end{equation}
Note that by introducing the auxiliary matrix $\bV$, the task of optimizing~\eqref{eq:S_udpate_SV} w.r.t. $\bS$ can be done within the continuous domain. However, due to the binary restrictions on $\bV$, solving~\eqref{eq:S_udpate_SV} is still a challenging problem. 

In this work, we propose to tackle~\eqref{eq:S_udpate_SV} by using the $\ell_1$ penalty method~\cite{wright1999numerical}. Specifically, by incorporating the coupling constraint $\bS = \bV$ into the cost function of~\eqref{eq:S_udpate_SV}, in the context of $\ell_1$ penalty method, the penalty problem can be written as follows
\begin{equation}
    \label{eq:S_udpate_SV_penalty}
    \begin{aligned}
    & \min_{\bS, \bV}  && \langle \bY, \bS \rangle + \rho \|\bS-\bV\|_1, \\
    & \text{subject to} && 0 \le \bS_{i,j} \le 1 \;\; \forall i,j,  \\
    & && \bS \in \bOmega, \;\; \bV \in \{0,1\}^{k\times N}, \\    
    \end{aligned}
\end{equation}
where the notation $\|.\|_1$ of a matrix $\bX$ is defined as  $\|\bX\|_1 = \sum_{i,j} |\bX_{i,j}|$ and $\rho > 0$ is the penalty parameter. Intuitively, by minimizing the sum of $\sum_{i,j}|\bS_{i,j} - \bV_{i,j}|$ with a penalty parameter $\rho$, the element-wise differences between the two matrices $\bS$ and $\bV$ are penalized during the optimization process, where the penalization strength is controlled by $\rho$. 

Furthermore, observe that the equality constraint $\bS_{i,j} = \bV_{i,j}$ can also be represented by two inequality constraints, i.e., $\bS_{i,j} \ge \bV_{i,j}$ and $\bV_{i,j} \ge \bS_{i,j}$. This allows us to write the second term containing the $\ell_1$ norm in~\eqref{eq:S_udpate_SV_penalty} as the sum of two separate terms. Specifically, let $[x]^- = \max\{0, -x\}$, and introduce two non-negative matrices $\brho^+ \in \bbR^{k \times N}$ and $\brho^- \in \bbR^{k\times N}$ that store the penalty parameters, the penalty problem~\eqref{eq:S_udpate_SV_penalty} becomes
\begin{equation}
    \label{eq:S_udpate_SV_penalty_2}
    \begin{aligned}
    & \min_{\bS, \bV}  && \langle \bY, \bS \rangle + \sum_{i,j}\brho^+_{i,j} [\bS_{i,j}-\bV_{i,j}]^- + \sum_{i,j}\brho_{i,j}^- [\bV_{i,j} - \bS_{i,j}]^-,  \\
    & \text{subject to} && 0 \le \bS_{i,j} \le 1 \;\; \forall i,j,  \\
    & && \bS \in \bOmega, \;\; \bV \in \{0,1\}^{k\times N}. \\    
    \end{aligned}
\end{equation}
Note that instead of using one penalty parameter $\rho$ as in~\eqref{eq:S_udpate_SV_penalty}, we associate each matrix index $(i,j)$ with two penalty parameters $\brho^+_{i,j} > 0$ and $\brho^-_{i,j} > 0$ that correspond to the constraints $\bS_{i,j} \ge \bV_{i,j}$ and $\bV_{i,j} \ge \bS_{i,j}$, respectively.

We are now ready to introduce the alternating optimization scheme to solve the problem \eqref{eq:S_udpate_SV_penalty_2}. The algorithm consists of iteratively updating $\bS$ and $\bV$ until convergence. Particularly, let $t$ denote the iteration, the update steps for $\bS$ and $\bV$ can be formulated as follows

\subsubsection{S update step} With $\bV$ fixed to the value at iteration $t$, updating $\bS$ at iteration $t+1$ amounts to solving a linear programming (LP) problem.  Specifically, by introducing two non-negative matrices $\bgamma^+ \in \bbR^{k \times N}$ and $\bgamma^- \in \bbR^{k \times N}$, the LP problem to update $\bS$ can be written as follows
\begin{subequations}
    \label{eq:S_udpate_SV_penalty_S}
    \begin{align}
    & \bS^{(t+1)} = \arg\min_{\bS, \bgamma}  && \langle \bY, \bS \rangle + \sum_{i,j}\brho^+_{i,j} \bgamma^+_{i,j} + \sum_{i,j}\brho^-_{i,j} \bgamma^-_{i,j} \label{eq:S_udpate_SV_penalty_S:obj}, \\
    &~~~~~~~~~~~~\text{subject to} && \bV^{(t)}_{i,j}-\bS_{i,j} \le \bgamma^+_{i,j} \; \forall i,j \label{eq:S_udpate_SV_penalty_S:gamma1}, \\    
    & &&  \bS_{i,j} - \bV^{(t)}_{i,j} \le \bgamma^-_{i,j} \;\; \forall i,j \label{eq:S_udpate_SV_penalty_S:gamma2},\\
    & &&  \bgamma^+_{i,j}, \bgamma^-_{i,j} \ge 0 \;\;\ \forall i,j, \\
    & && 0 \le \bS_{i,j} \le 1 \;\; \forall i,j, \;\;\;\bS \in \bOmega. 
    \end{align}
\end{subequations}
Note that the purpose of introducing $\bgamma^+$ and $\bgamma^-$ is to eliminate the $[.]^-$ operator in the objective function of~\eqref{eq:S_udpate_SV_penalty_2} -- based on the observation that $[x]^- \ge 0 \;\forall x$. Therefore, the problem~\eqref{eq:S_udpate_SV_penalty_2} and~\eqref{eq:S_udpate_SV_penalty_S} with fixed $\bV$ are equivalent. As~\eqref{eq:S_udpate_SV_penalty_S} is a convex LP problem, it can be solved efficiently using any off-the-shelf solver.

\subsubsection{V update step} After $\bS$ is updated by solving the LP problem~\eqref{eq:S_udpate_SV_penalty_S}, starting from~\eqref{eq:S_udpate_SV_penalty_2}, the task of updating $\bV$ with fixed $\bS$ can be written as 
\begin{equation}
    \label{eq:S_udpate_SV_penalty_V}
    \begin{aligned}
    & \bV^{(t+1)} = \;\arg\min_{\bV}\;  && \sum_{i,j}\brho^+_{i,j} [\bS^{(t+1)}_{i,j}-\bV_{i,j}]^- + \sum_{i,j}\brho_{i,j}^- [\bV_{i,j} - \bS^{(t+1)}_{i,j}]^-,  \\
    &~~~~~~~~~~~~~~\text{subject to} &&  \bV \in \{0,1\}^{k\times N}.  \\        
    \end{aligned}
\end{equation}
Due to the fact that the objective function of~\eqref{eq:S_udpate_SV_penalty_V} consists of non-negative terms, solving for $\bV$ can be done element-wise. Moreover, as each element $\bV_{i,j} \in \{0,1\}$, it can be updated by choosing the value that results in the smaller objective value
\begin{equation}
    \label{eq:S_update_SV_V_closedform}
    \bV^{(t+1)}_{i,j} = \begin{cases} 
        0, \;\;\; \text{if} \;\; \brho^-_{i,j}\bS^{(t+1)}_{i,j} \le  \brho_{i,j}^+ \left(1 - \bS^{(t+1)}_{i,j}\right),  \\
        1, \;\;\; \text{otherwise.}
    \end{cases}
\end{equation}

\subsubsection{Updating penalty parameters}
From~\eqref{eq:S_update_SV_V_closedform}, it can be seen that for a fixed value of $\bS^{(t+1)}_{i,j}$, the penalty parameters $\brho^+_{i,j}$ and $\brho^-_{i,j}$ control the weights for updating $\bV^{(t+1)}_{i,j}$. Particularly, if $\brho^+_{i,j}$ is much larger than $\brho^-_{i,j}$, the value of $0$ is more favorable for $\bV^{(t+1)}_{i,j}$ (since $ \brho_{i,j}^+ (1 - \bS^{(t+1)}_{i,j}) \gg \brho^-_{i,j}\bS^{(t+1)}_{i,j})$ , and vice versa. Thus, in other to prevent early convergence to a bad local minima, if $\bV^{(t+1)}_{i,j} = 0$ at iteration $t+1$, we increase $\brho^-_{i,j}$ to give $\bV_{i,j}$ chances to be assigned with the value of $1$ in the later iterations. Similar argument can be applied for the case of $\bV^{(t+1)}_{i,j}=1$. The penalty parameters, therefore, are updated as follows
\begin{equation}
    \label{eq:rho_update}
    \brho^{-(t+1)}_{i,j} =  \begin{cases}
        \kappa\brho^{-(t)}_{i,j}, \;\; \text{if}\;\; \bV^{(t+1)}_{i,j} = 0, \\
        \brho^{-(t)}_{i,j}, \;\;\; \text{otherwise},    
    \end{cases} \text {and} \;\;
    \brho^{+(t+1)}_{i,j} =  \begin{cases}
        \kappa\brho^{+(t)}_{i,j}, \;\; \text{if}\;\; \bV^{(t+1)}_{i,j} = 1, \\
        \brho^{+(t)}_{i,j}, \;\;\; \text{otherwise},    
    \end{cases}
\end{equation}
where $\kappa$ is a positive number that controls the increase rate. Note that besides controlling the updating of $\bV$ in~\eqref{eq:S_update_SV_V_closedform}, the $\brho^+$ and $\brho^-$ parameters also affect the solution of~\eqref{eq:S_udpate_SV_penalty_S}. Specifically, by gradually increasing $\brho^+$ and $\brho^-$ as in~\eqref{eq:rho_update}, the weights for the second and third terms in the objective function~\eqref{eq:S_udpate_SV_penalty_S:obj} become higher in the later iterations, forcing~\eqref{eq:S_udpate_SV_penalty_S} to drive $\bS$ to integrality.

\section{Main algorithm}

\begin{algorithm}[ht]\centering
    \caption{Binary Optimization Based Constrained K-Means (BCKM)}
    \label{alg:bckmeans}                         
    \begin{algorithmic}[1]                   
        \REQUIRE Input data $\bX$, number of clusters \ttnclusters, convergence threshold $\epsilon_c$, \ttmaxiter, initial assignment $\bS^{(0)}$, initial centroids $\bC^{(0)}$ 
        \STATE t $\leftarrow 1$; \;\; 
        \WHILE{t $<$ \ttmaxiter}                        
        \STATE $\bC^{(t)} \leftarrow$ $\arg\min_{\bC} \|\bX - \bC \bS^{(t-1)}\|^2_F$          \label{alg:bckmeans:centroid}
        \STATE $\bS^{(t)} \leftarrow$ UpdateClusterAssignment($\bX, \bC^{(t)}$, $\bS^{(t-1)}$) /*Alg.\ref{alg:update_assigment} */  \label{alg:bckmeans:assignment}
            \IF{$\|\bC^{(t)} - \bC^{(t-1)}\|^2_F \le \epsilon$}  
                \STATE break
            \ENDIF
            \STATE $t \leftarrow  t + 1$
        \ENDWHILE
        \RETURN Cluster centroids $\bC$, cluster assignment matrix $\bS$
    \end{algorithmic}
\end{algorithm}

\begin{algorithm}[ht]\centering
    \caption{Update Cluster Assignment}
    \label{alg:update_assigment}                         
    \begin{algorithmic}[1]                   
        \REQUIRE Data  matrix $\bX$, set of centroids $\bC$, $\bS^{(0)}$, initial penalty parameter $\rho_0$, penalty increase rate $\kappa$, convergence threshold $\epsilon_s$,  \ttmaxiter
        \STATE t $\leftarrow 1$; \;\; $\bY_{i,j} \leftarrow \|\bc_i - \bx_j\|^2_2$
        \STATE $\brho^+ \leftarrow \rho_0 \mathbf{1}^{k \times N}$; $\brho^- \leftarrow \rho_0 \mathbf{1}^{k \times N}$                  
        \STATE $\bV^{(0)} \leftarrow [\bS^{(0)}]$ \;\;/*$[.]$ denotes rounding to nearest integer*/
        \WHILE{t $<$ \ttmaxiter}                        
            \STATE Update $\bS^{(t)}$ using~\eqref{eq:S_udpate_SV_penalty_S} with $\bV$ fixed to $\bV^{(t-1)}$            
            \STATE Update $\bV^{(t)}$ using~\eqref{eq:S_update_SV_V_closedform} with $\bS$ fixed to $\bS^{(t)}$
            \STATE Update $\brho^{+(t)}$ and $\brho^{-(t)}$ using~\eqref{eq:rho_update}
            \IF {$\| \bS^{(t)}- \bS^{(t-1)} \|^2_F + \|\bV^{(t)}- \bV^{(t-1)}  \|^2_F \le \epsilon_s $ }
                \STATE break
            \ENDIF
        \ENDWHILE
        \RETURN $\bS$ 
    \end{algorithmic}
\end{algorithm}

Based on the discussions in the previous sections, Algorithm~\ref{alg:bckmeans} summarizes our main approach for solving the constrained K-Means problem~\eqref{eq:constrained_k_means}. The algorithm alternates between updating the centroids~ (Line \ref{alg:bckmeans:centroid}, ~Algorithm~\ref{alg:bckmeans}) and updating the cluster assignments (Algorithm~\ref{alg:update_assigment}). Note that in Line~\ref{alg:bckmeans:assignment} of Algorithm~\ref{alg:bckmeans}, the current value of $\bS$ is supplied to Algorithm~\ref{alg:update_assigment} for initialization. In Algorithm~\ref{alg:bckmeans}, $\epsilon_c$ is the convergence threshold of the clusters, i.e., we stop the algorithm if the Frobenius norm of two consecutive centroid matrices is less than $\epsilon_c$. Similarly, the parameter $\epsilon_s$ in Algorithm~\ref{alg:update_assigment} determines the stopping condition for two consecutive set of variables $(\bS,\bV)$. In Algorithm~\ref{alg:update_assigment}, each elements of the penalty parameter matrices $\brho^+$ and $\brho^-$ are initialized to the same value $\rho_0$ (note that $\mathbf{1}^{k\times N}$ denotes a matrix of size $k\times N$ with all elements equal to $1$). 

\section{Experiments}
In this section, we evaluate the performance of our proposed algorithm (BCKM -- Binary Optimization based Constrained K-Means) on synthetic and real datasets and compare BCKM with several popular approaches. Among a large body of works on constrained K-Means, we only select some commonly used and state-of-the-art representatives to benchmark our algorithm against, including:
The conventional K-Means clustering algorithm (KM)~\cite{macqueen1967some}; Hierarchical K-Means (HKM)~\cite{johnson1967hierarchical}; Constrained K-Means with Background Knowledge (COP-KM)~\cite{wagstaff2001constrained}; Balanced K-Means (BKM)~\cite{malinen2014balanced}; Constrained K-Means (CKM)~\cite{bradley2000constrained}; Constrained K-Means with Spectral Clustering (CKSC)~\cite{li2009constrained}; Balanced Clustering with Least Square Regression (BCLR)~\cite{liu2017balanced}.

All experiments are executed on a standard Ubuntu desktop machine with 4.2GHz CPU and 32 GB of RAM. We implement our proposed method in Python. All the runs are initialized with standard K-Means. The $\rho_0$ parameter is set to the starting value of $0.5$ and is increased by a rate of $\kappa = 1.1$ for all experiments. For KM and HKM, we employ the implementation provided by the Scikit-learn library~\cite{pedregosa2011scikit} with $10$ random initializations and the maximum number of iterations is set to $100$. For BCLR, we use the MATLAB code and parameters provided by the authors~\cite{liu2017balanced}, with maximum number of iterations set to $2000$.   
We use our own Python implementation of Constrained K-Means~\cite{bradley2000constrained} and Balanced K-Means~\cite{malinen2014balanced}. To measure the performance of the algorithms, we report the Normalized Mutual Information (NMI), which is a commonly used metric for clustering problems. Additionally, in order to evaluate the efficiency of our method compared to other approaches, we also report the run time (in seconds) for all the experiments.
\vspace{-0.2cm}
\subsection{Balanced clustering on synthetic data}
In this experiment, we test the performance of the methods on the task of balanced clustering with must-link and cannot-link constraints on high dimensional data. Note that our algorithm, based on the formulation~\eqref{eq:constrained_k_means_01}, is capable of handling different bounds on cluster sizes. However, we conduct experiments on balanced clustering to provide a fair comparison for algorithms that can only handle balance constraints~\cite{liu2017balanced}. 
We randomly generate $k$ clusters, where each cluster contains $n$ data points. We choose $N = kn \approx 500$ data points, and each data point $\bx_i$ belongs the space of $\bbR^{d}$ (with $d=512$).  
To generate the set containing $k$ cluster centers $\{\mu_i\}_{i=1}^k$, we uniformly sample $k$ points in the hyperbox $[-1,1]^d$. For each $i$-th cluster, its members are generated by randomly drawing $n$ points from a Gaussian distribution with mean $\mu_i$ and covariance matrix of $\sigma\bI$ ($\bI$ is the identity matrix). 
To achieve balanced clustering, we set the lower bounds on the cluster sizes for all clusters to be $n$.  Besides, within each cluster, $20\%$ of the points are randomly sampled to generate must-link and cannot-link constraints.  

\begin{figure}
    \centering
    \includegraphics[width = 0.45\textwidth]{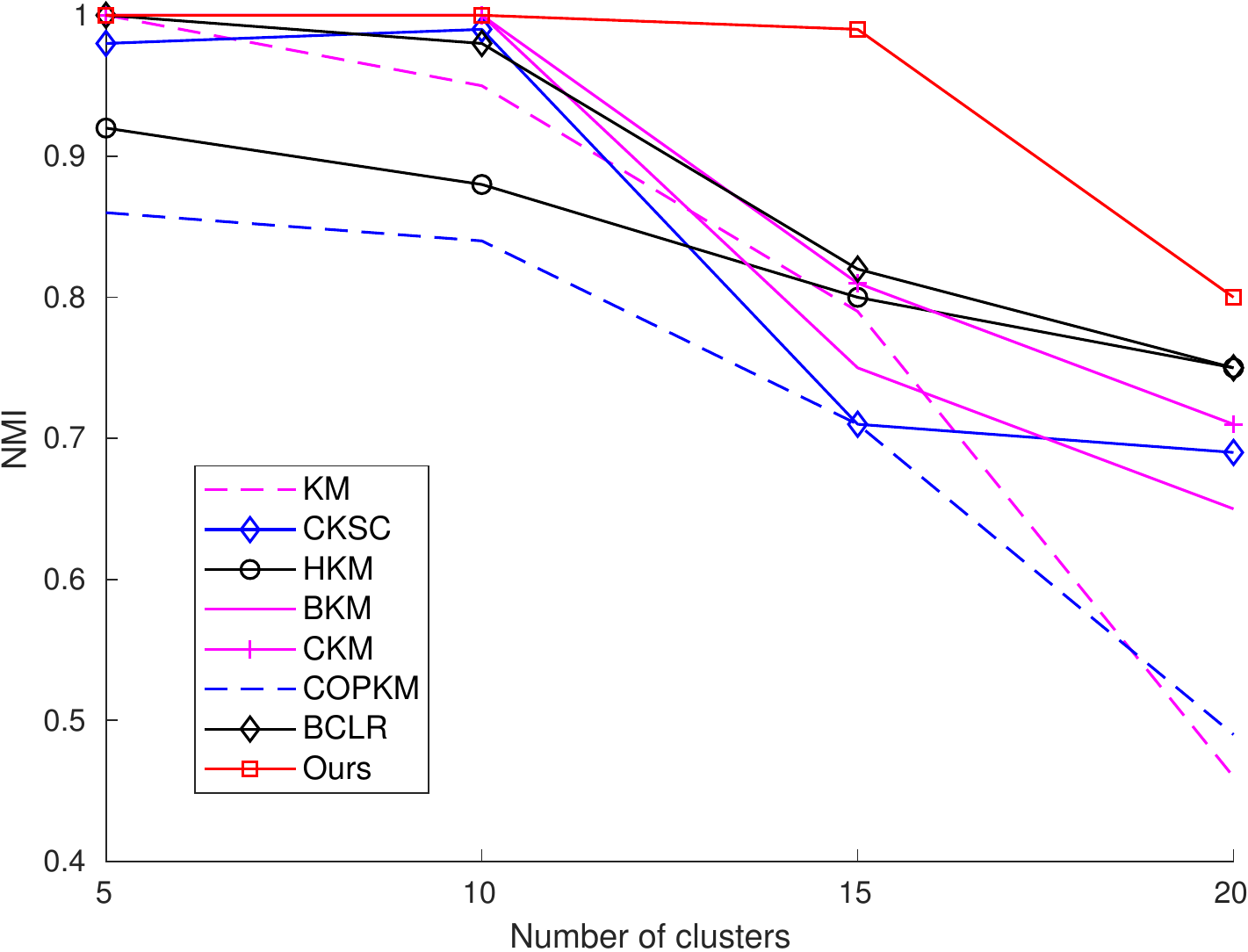}
    \includegraphics[width = 0.45\textwidth]{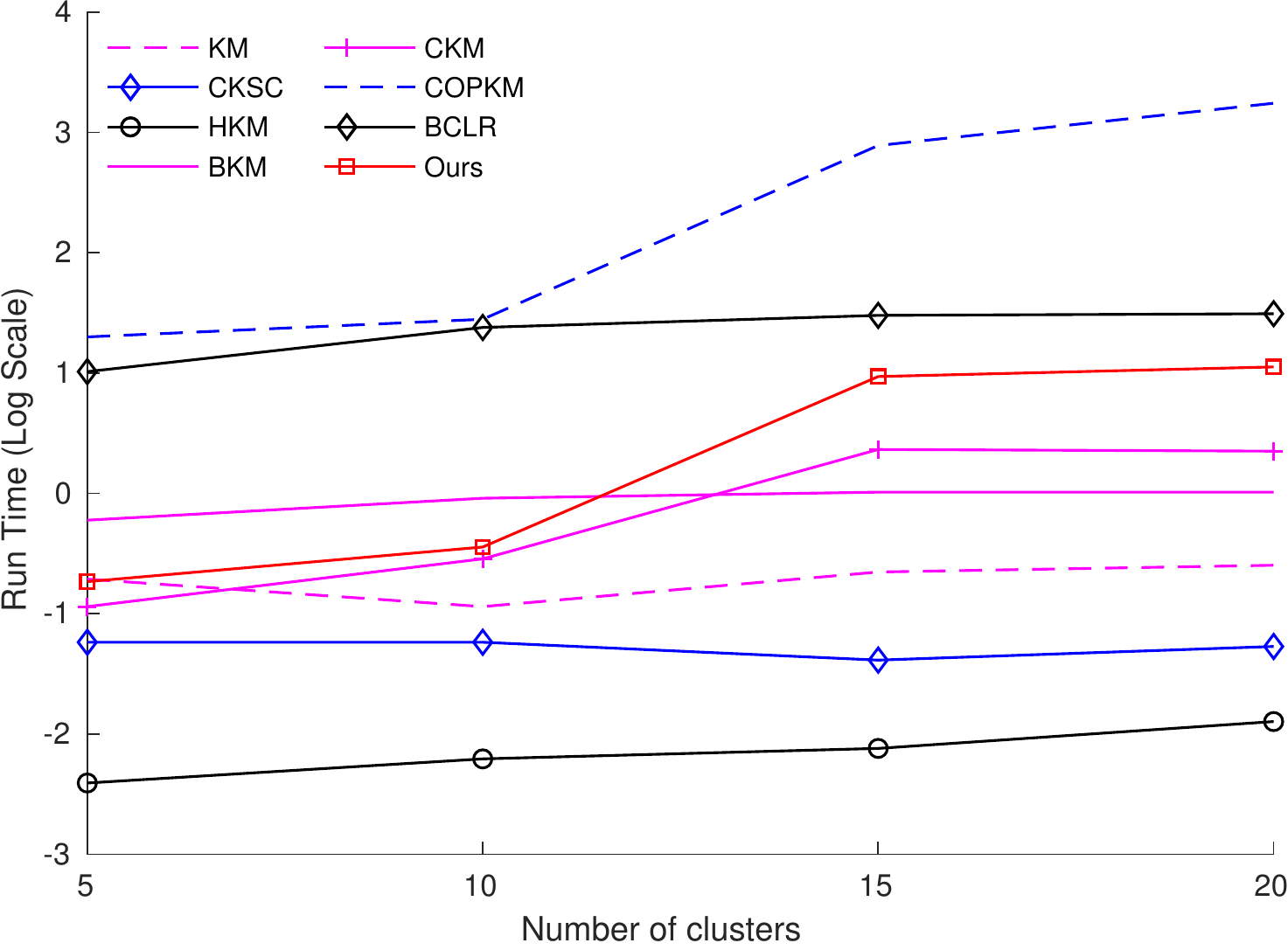}\\
    \includegraphics[width = 0.45\textwidth]{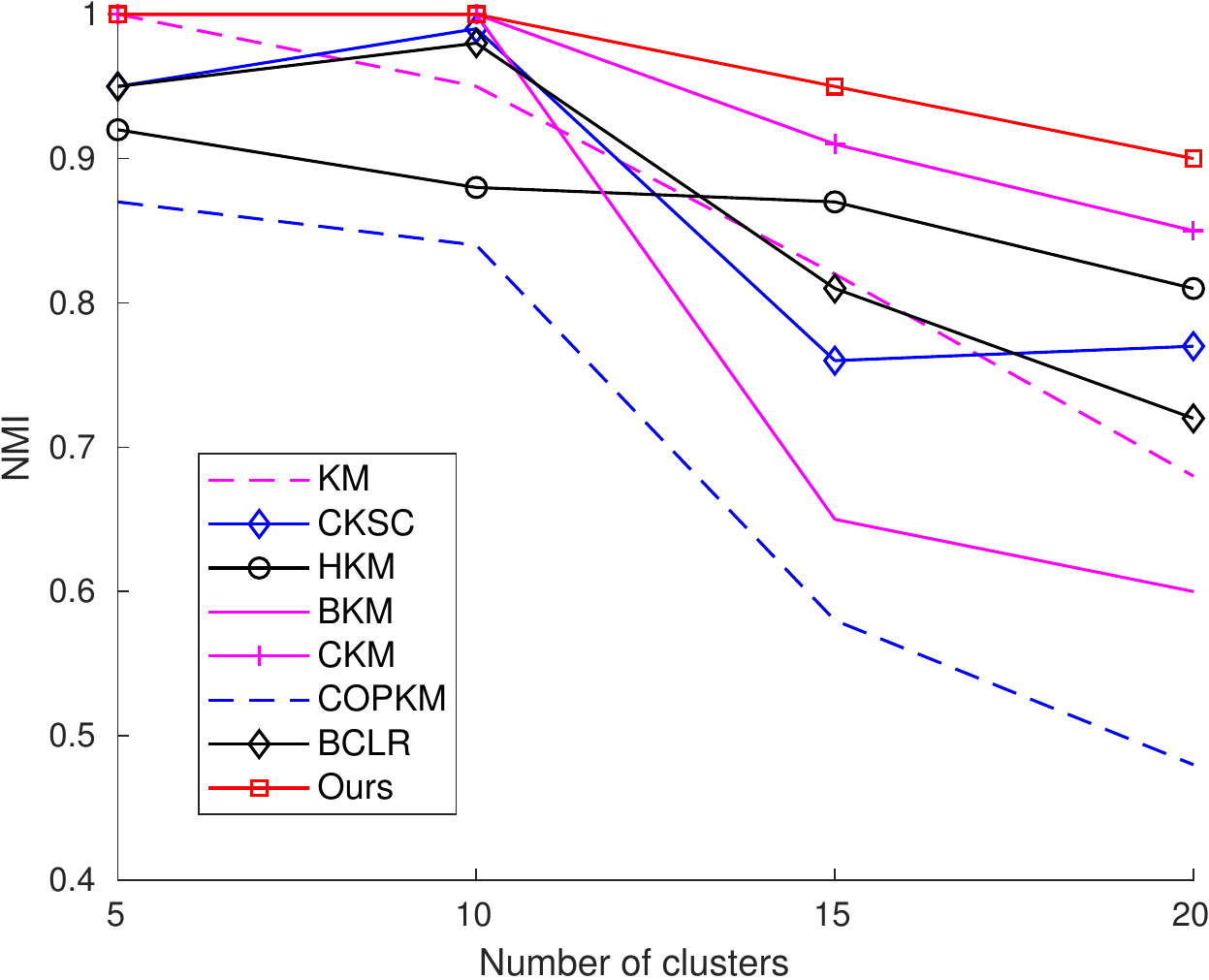}
    \includegraphics[width = 0.45\textwidth]{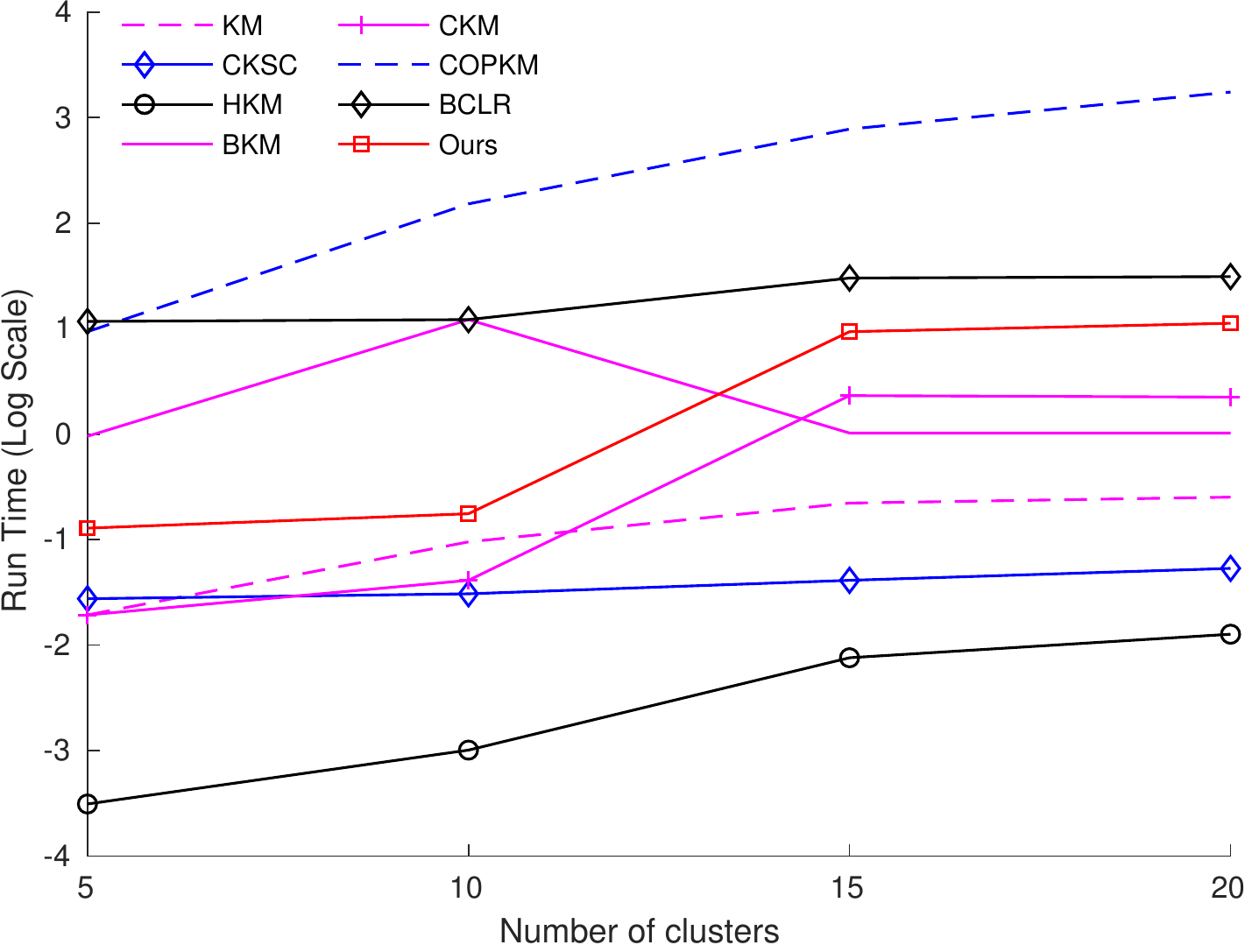}\\
    \includegraphics[width = 0.45\textwidth]{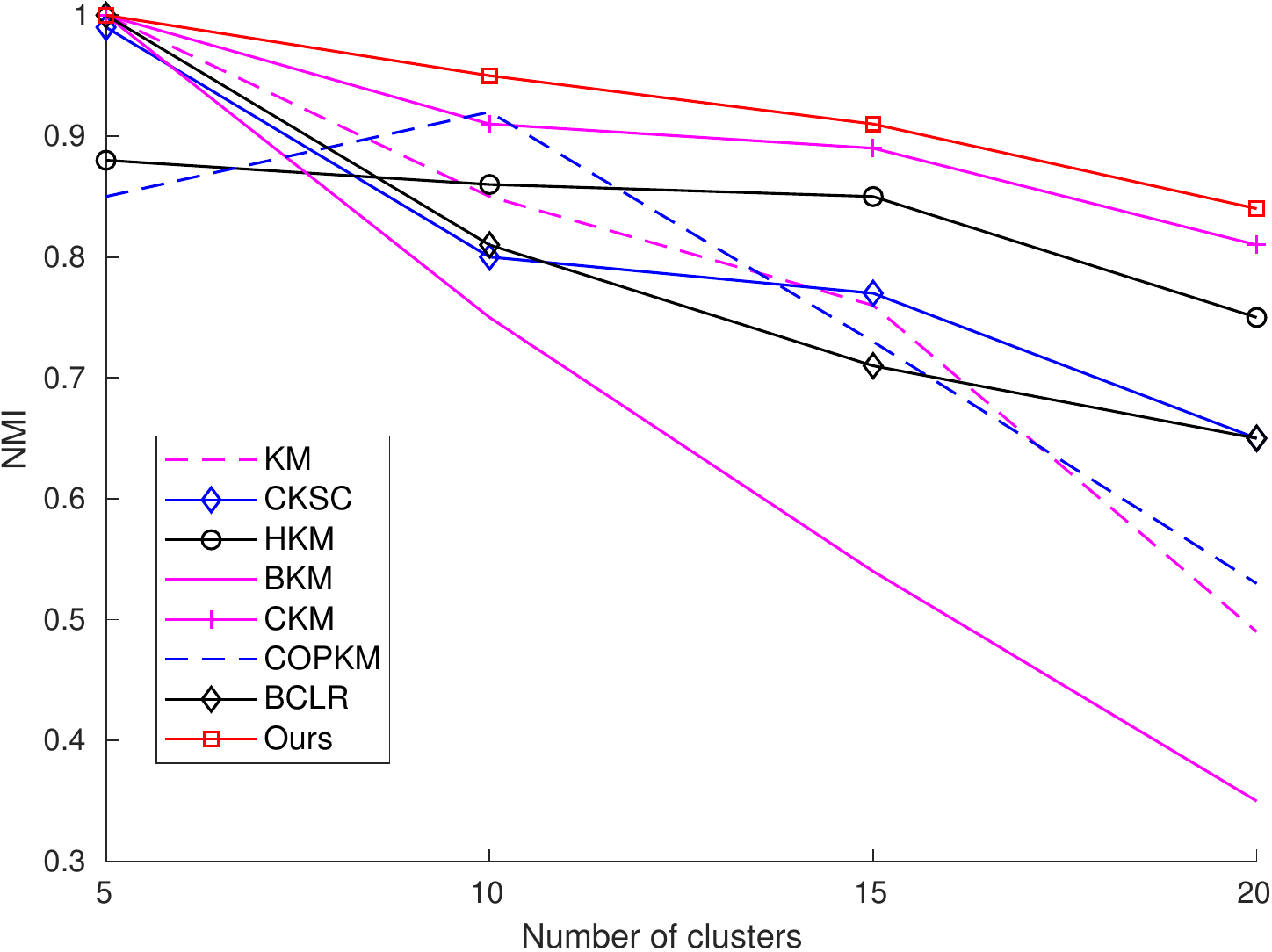} 
    \includegraphics[width = 0.45\textwidth]{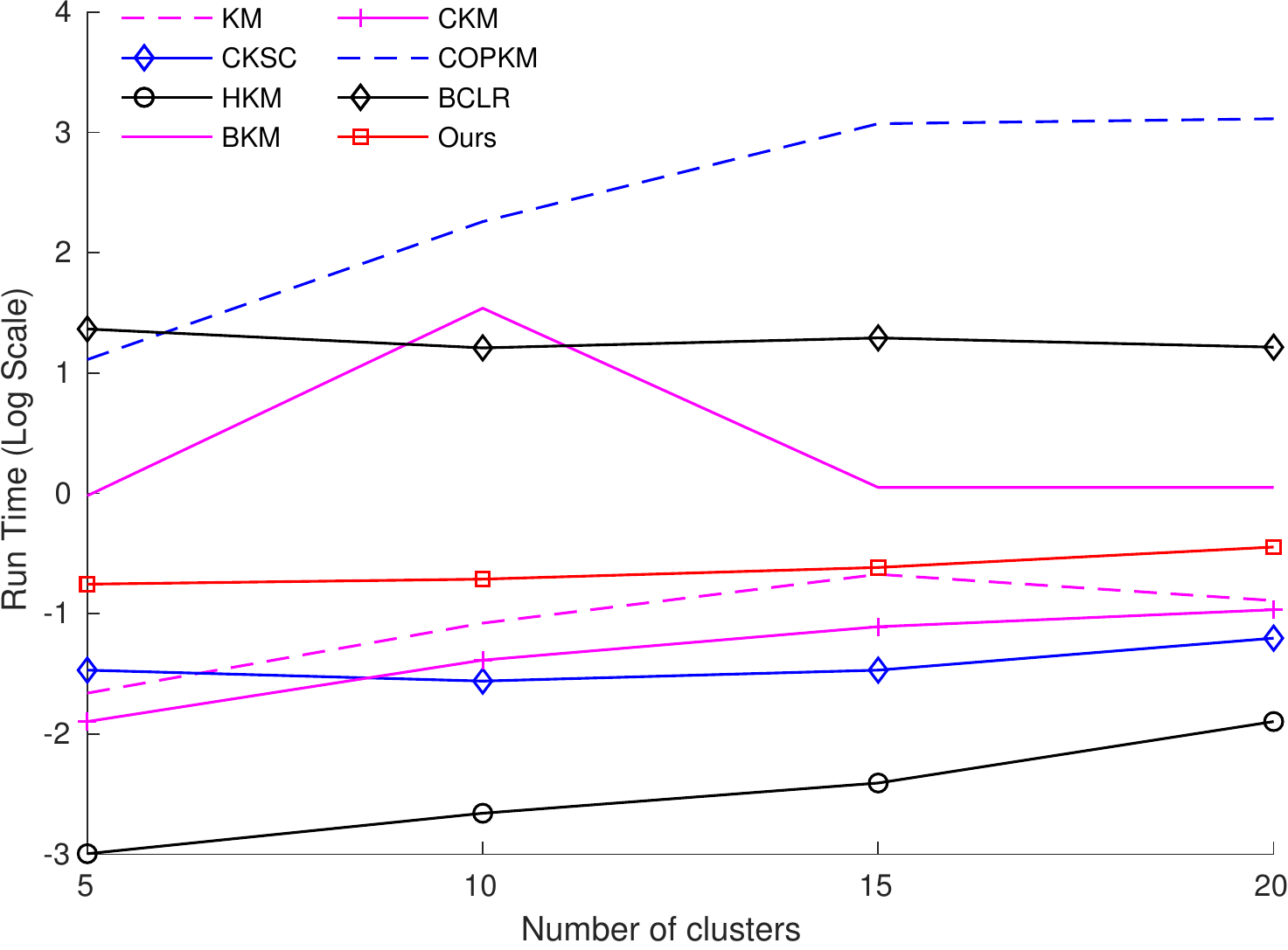}
    \caption{NMI (left column) and run time (right) for different values of clusters with $N=500$ points and $d=512$. First row: NMI and run time for $\sigma = 0.1$. Second row: NMI and run time for $\sigma = 0.5$. Last row: NMI and run time for $\sigma = 0.7$.}
    \label{fig:synthetic_results}
\end{figure}

Figure~\ref{fig:synthetic_results} shows experiment results for $\sigma = 0.1$ (top row), $\sigma = 0.5$ (second row) and $\sigma = 0.7$ (bottom row), respectively. On each row of Figure~\ref{fig:synthetic_results}, we plot the NMI (left) and run time (right) for all the methods.
As can be observed from this figure, with small values of $k$, all methods provide relatively good clustering results. As $k$ increases, however, the performances of all methods also degrade (with lower NMI). Among them, our proposed method provides the best NMI result due to its ability to strictly enforce the constraints. Note that although the linking constraints are also added to the LP formulation in~\cite{bradley2000constrained}, our method is able to achieve much higher NMI due to its ability to find good binary solutions, while our runtime is only slightly higher compared to that of~\cite{bradley2000constrained}. Observe that as the value of $\sigma$ increases, the performance of the methods also degrade, but ours is able to achieve the best NMI compared to others as the constraints are properly enforced.

\begin{figure}
    \centering
    \includegraphics[width = 0.31\textwidth]{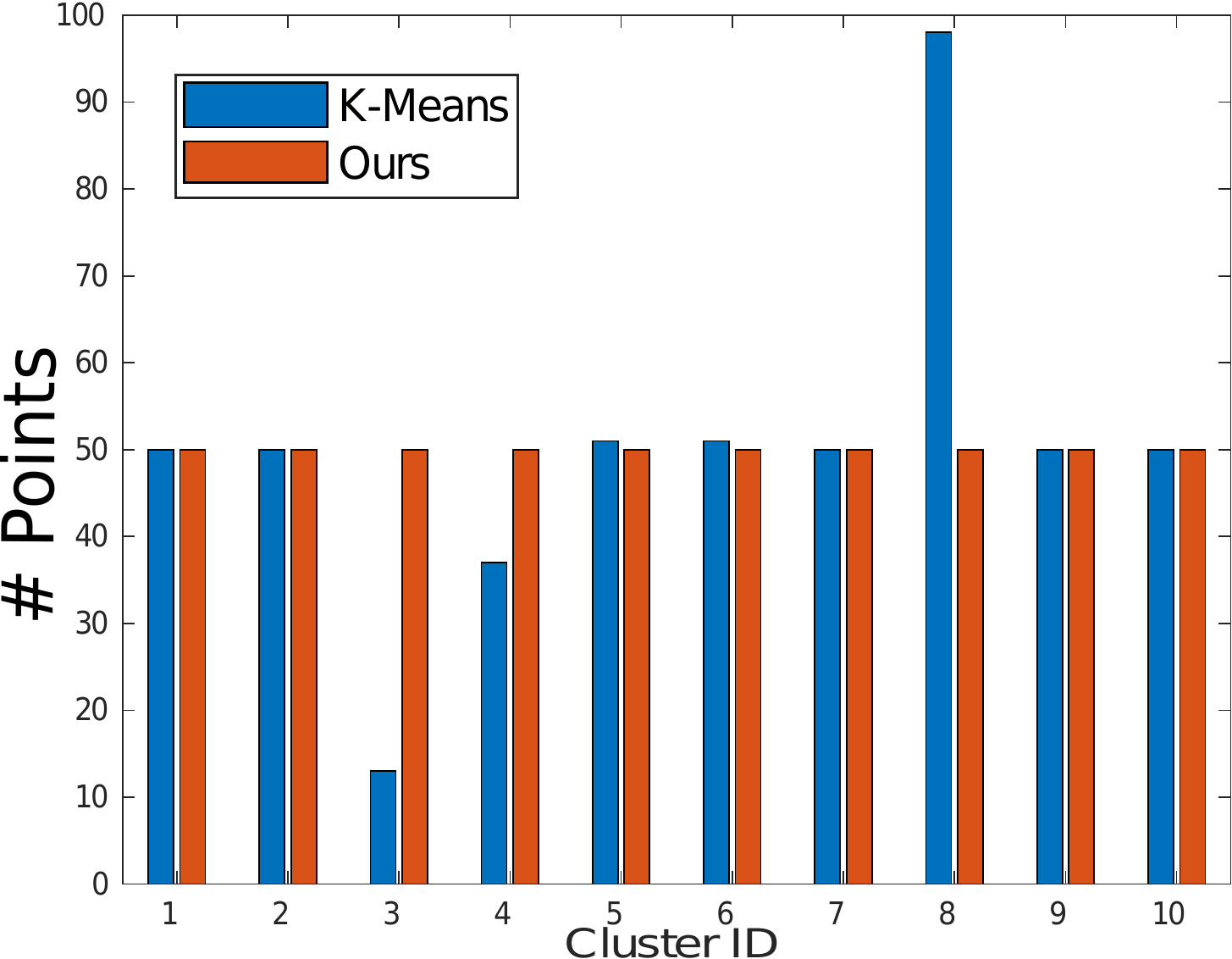}
    \includegraphics[width = 0.31\textwidth]{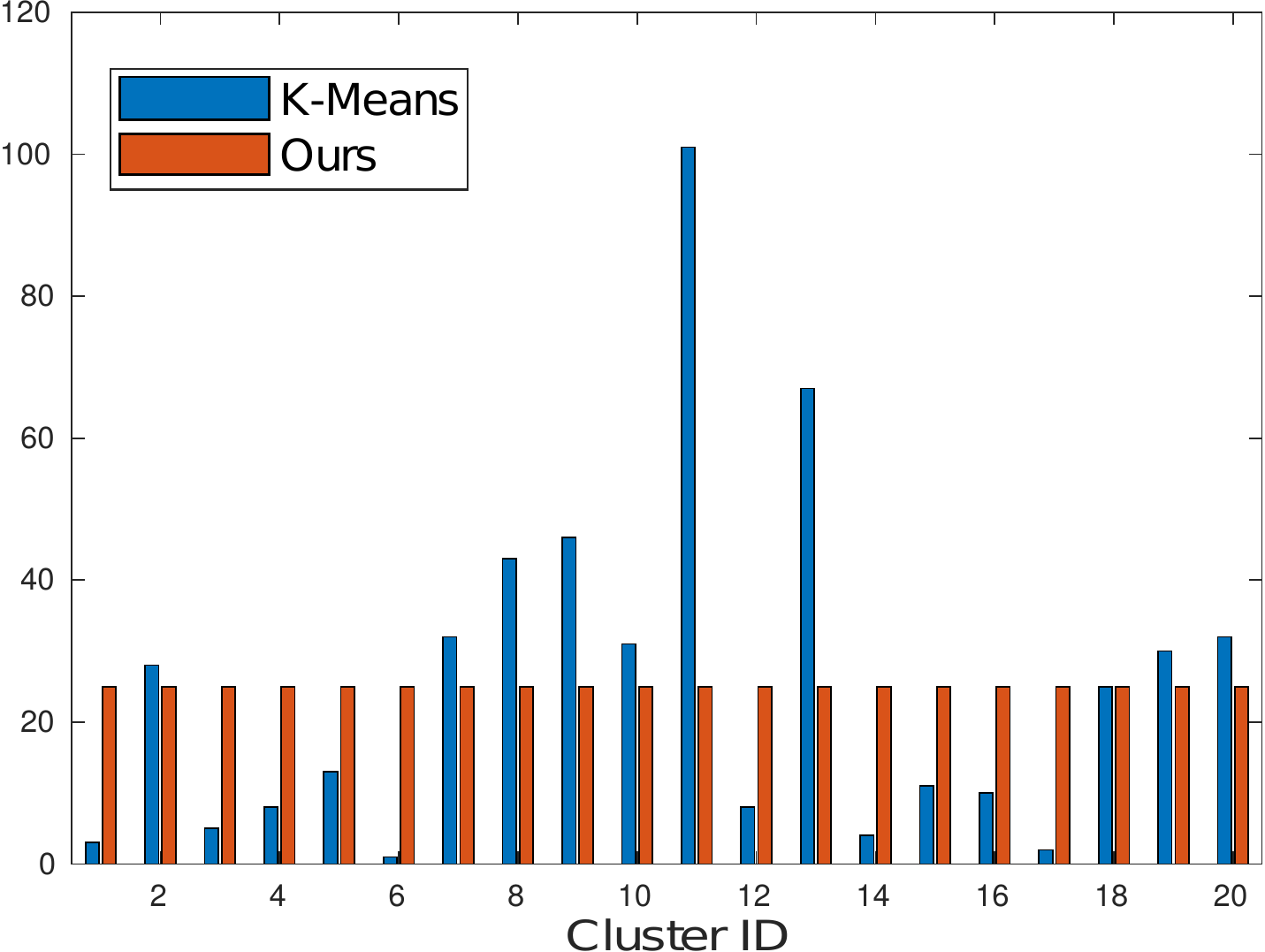}
    \includegraphics[width = 0.31\textwidth]{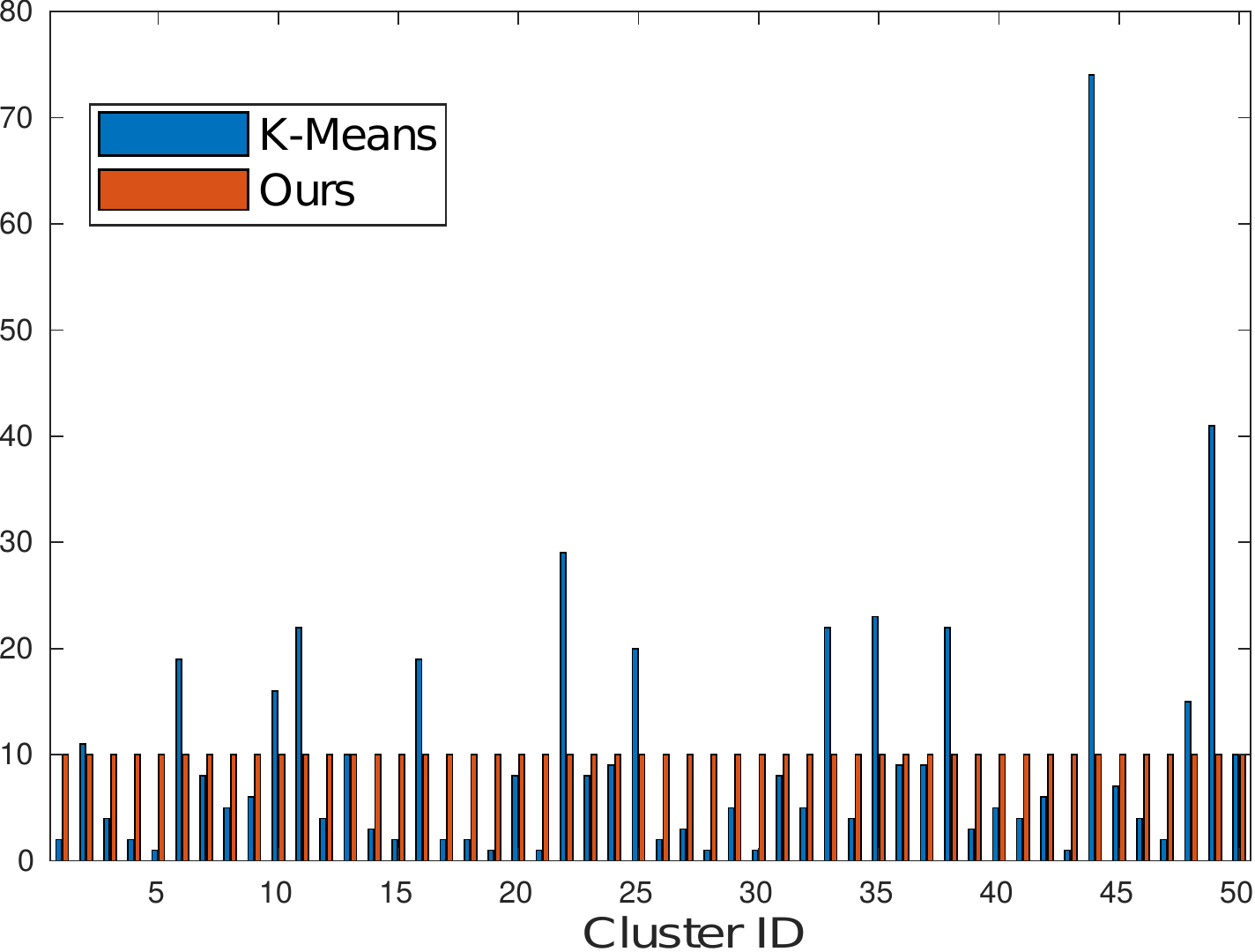}
    \caption{Cluster distribution of K-Means and our method for 3 different values of $k={10, 20, 50}$ (from left to right). }
    \label{fig:synthetic_cluster_distribution}
\end{figure}

To demonstrate the ability to provide balanced clusters, we plot in Figure~\ref{fig:synthetic_cluster_distribution} the number of points distributed into each cluster by K-Means and our method for three different values of $k$ (we use K-Means to initialize our method). Observe that as $k$ increases, the clusters provided by K-Means becomes highly unbalanced. With such unbalanced initializations, however, our method is able to refine the initial solutions and return balanced clusters, while the must-link and cannot-link constraints are also enforced to provide better NMI compared to other approaches.

\subsection{Clustering on real datasets}
Besides testing with synthetic data, we also conduct experiments on some real datasets that are often employed to benchmark clustering algorithms, including MNIST~\cite{lecun1998gradient}, ORL Face Dataset\footnote{http://www.cl.cam.ac.uk/research/dtg/attarchive/facedatabase.html}, UMIST Face Datset\footnote{https://www.sheffield.ac.uk/eee/research/iel/research/face}, Yale Dataset\footnote{http://vision.ucsd.edu/content/yale-face-database} and YaleB Dataset\footnote{http://vision.ucsd.edu/content/extended-yale-face-database-b-b}.   
For each dataset, we randomly sample data points from 10 to 15 clusters, and each cluster contains the same number of instances (for the task of balanced clustering). Within each cluster, we randomly select 20\% of the points to enforce the must-link and cannot-link constraints. For large datasets, we repeat the experiment with different random subsets of data (the subset indexes are shown by the number in the parentheses). The NMI and the run time of the methods are shown in Table.~\ref{table:real}. Similar to the case of synthetic data, throughout most of the experiments, our method provides better NMI compared to other approaches due to its ability to provide assignments that satisfy the constraints. Note that in some cases, the NMI results provided by BCLS and COP are very close to ours, while our run time is much faster. We also apply the must-link and cannot-link constraints to the LP formulation of CKM~\cite{bradley2000constrained}, yet we are able to achieve better NMI. This demonstrates that our binary optimization technique provides better solutions compared to the LP relaxation approach.

\begin{table}[ht]
\centering
\resizebox{0.95\textwidth}{!}{
    \begin{tabular}{c|c|c c c c c c c c}
    \hline\hline
    Datasets               &      & KM\cite{macqueen1967some}   & CKSC\cite{li2009constrained}     & HKC\cite{johnson1967hierarchical}   & BKM\cite{malinen2014balanced}  & CKM\cite{bradley2000constrained}  & BCLS\cite{liu2017balanced} & COP\cite{wagstaff2001constrained}  & Ours          \\ \hline
    \multirow{2}{*}{MNIST (1)} & NMI  & 0.49 & 0.62 & 0.57 & 0.51 & 0.68 & 0.62 & 0.68 & \textbf{0.69} \\ 
                        & Time & 0.63 & 0.15  & 0.11 & 0.52 & 0.40 & 2.52 & 5.54 & 0.49              \\ \hline\hline

    \multirow{2}{*}{MNIST (2)} & NMI  & 0.52 & 0.55 & 0.54 & 0.52 & 0.58 & 0.50 & 0.48 & \textbf{0.60} \\ 
                            & Time & 0.77 & 0.12  & 0.14 & 0.47 & 0.33 & 3.36 & 4.07 & 0.40              \\ \hline\hline

    \multirow{2}{*}{MNIST (3)} & NMI  & 0.53 & 0.59 & 0.57 & 0.50 & 0.59 & 0.55 & 0.56 & \textbf{0.61} \\ 
                            & Time & 0.77 & 0.12  & 0.14 & 0.47 & 0.45 & 2.19 & 4.52 & 0.54              \\ \hline\hline

    \multirow{2}{*}{Yale (1)} & NMI  & 0.51 & 0.54 & 0.59 & 0.51 & 0.67 & 0.55 & 0.61 & \textbf{0.73} \\ 
                            & Time & 0.34 & 0.09  & 0.15 & 0.23 & 0.24 & 4.13 & 2.13 & 0.30              \\ \hline\hline

    \multirow{2}{*}{Yale (2)} & NMI  & 0.54 & 0.52 & 0.51 & 0.50 & 0.64 & 0.68 & 0.56 & \textbf{0.72} \\ 
                            & Time & 0.34 & 0.17  & 0.13 & 0.27 & 0.33 & 3.21 & 3.11 & 0.39              \\ \hline\hline

    \multirow{2}{*}{YaleB (1)} & NMI  & 0.55 & 0.62 & 0.59 & 0.48 & 0.68 & 0.60 & 0.64 & \textbf{0.71} \\ 
                            & Time & 0.34 & 0.11  & 0.21 & 0.31 & 0.19 & 1.95 & 2.48 & 0.22              \\ \hline\hline

    \multirow{2}{*}{YaleB (2)} & NMI  & 0.52 & 0.53 & 0.54 & 0.50 & 0.61 & 0.55 & 0.54 & \textbf{0.66} \\ 
                            & Time & 0.34 & 0.09  & 0.08 & 0.43 & 0.20 & 2.39 & 2.40 & 0.25              \\ \hline\hline

    \multirow{2}{*}{YaleB (3)} & NMI  & 0.51 & 0.61 & 0.60 & 0.48 & 0.67 & \textbf{0.72} & 0.69 & \textbf{0.72} \\ 
                            & Time & 0.34 & 0.21  & 0.15 & 0.20 & 0.35 & 4.12 & 2.65 & 0.37              \\ \hline\hline

    \multirow{2}{*}{ORL (1)} & NMI  & 0.67 & 0.70 & 0.75 & 0.64 & 0.79 & 0.75 & \textbf{0.85} & {0.81} \\ 
                            & Time & 0.38 & 0.15  & 0.17 & 0.29 & 0.41 & 2.35 & 3.51 & 0.54             \\ \hline\hline

    \multirow{2}{*}{ORL (2)} & NMI  & 0.61 & 0.71 & 0.69 & 0.61 & 0.68 & 0.75 & 0.75 & \textbf{0.83} \\ 
                            & Time & 0.38 & 0.14  & 0.17 & 0.26 & 0.37 & 2.35 & 3.50 & 0.45             \\ \hline\hline

    \multirow{2}{*}{UMIST} & NMI  & 0.76 & \textbf{0.81} & 0.67 & 0.75 & 0.72 & 0.75 & 0.78 & \textbf{0.81} \\ 
                            & Time & 0.25 & 0.09  & 0.15 & 0.52 & 0.30 & 2.15 & 6.45 & 0.39              \\ \hline\hline


    \end{tabular}
}
\caption{Experiment results on real datasets. For each dataset, different subsets (indexed by the number in the parentheses) are sampled to run the experiments. Run time is in seconds.}
\label{table:real}
\end{table}
\vspace{-1cm}


\section{Conclusion}
In this work, we propose a binary optimization approach for the constrained K-Means clustering problem, in which different types of constraints can be simultaneously enforced. We then introduce a novel optimization technique to search for the solutions of the  problem in the binary domain, resulting in better solutions for the cluster assignment problem. Empirical results show that our method outperforms other heuristic or relaxation techniques while the increase in run time is negligible. The method proposed in this paper can be considered as a generic framework for constrained K-Means which can be embedded into different problems that require the use constrained clustering.
\bibliographystyle{splncs04}
\bibliography{quantization}

\begin{thebibliography}{10}
\providecommand{\url}[1]{\texttt{#1}}
\providecommand{\urlprefix}{URL }
\providecommand{\doi}[1]{https://doi.org/#1}

\bibitem{althoff2011balanced}
Althoff, T., Ulges, A., Dengel, A.: Balanced clustering for content-based image
  browsing. Series of the Gesellschaft fur Informatik pp. 27--30 (2011)

\bibitem{bertacco2007feasibility}
Bertacco, L., Fischetti, M., Lodi, A.: A feasibility pump heuristic for general
  mixed-integer problems. Discrete Optimization  \textbf{4}(1),  63--76 (2007)

\bibitem{bradley2000constrained}
Bradley, P., Bennett, K., Demiriz, A.: Constrained k-means clustering.
  Microsoft Research, Redmond pp.~1--8 (2000)

\bibitem{fard2018deep}
Fard, M.M., Thonet, T., Gaussier, E.: Deep $ k $-means: Jointly clustering with
  $ k $-means and learning representations. arXiv preprint arXiv:1806.10069
  (2018)

\bibitem{fischetti2005feasibility}
Fischetti, M., Glover, F., Lodi, A.: The feasibility pump. Mathematical
  Programming  \textbf{104}(1),  91--104 (2005)

\bibitem{ge2013optimized}
Ge, T., He, K., Ke, Q., Sun, J.: Optimized product quantization for approximate
  nearest neighbor search. In: Computer Vision and Pattern Recognition (CVPR),
  2013 IEEE Conference on. pp. 2946--2953. IEEE (2013)

\bibitem{geissler2017penalty}
Gei{\ss}ler, B., Morsi, A., Schewe, L., Schmidt, M.: Penalty alternating
  direction methods for mixed-integer optimization: A new view on feasibility
  pumps. SIAM Journal on Optimization  \textbf{27}(3),  1611--1636 (2017)

\bibitem{gersho2012vector}
Gersho, A., Gray, R.M.: Vector quantization and signal compression, vol.~159.
  Springer Science \& Business Media (2012)

\bibitem{jegou2011product}
Jegou, H., Douze, M., Schmid, C.: Product quantization for nearest neighbor
  search. IEEE transactions on pattern analysis and machine intelligence
  \textbf{33}(1),  117--128 (2011)

\bibitem{johnson1967hierarchical}
Johnson, S.C.: Hierarchical clustering schemes. Psychometrika  \textbf{32}(3),
  241--254 (1967)

\bibitem{kalantidis2014locally}
Kalantidis, Y., Avrithis, Y.: Locally optimized product quantization for
  approximate nearest neighbor search. In: Proceedings of the IEEE Conference
  on Computer Vision and Pattern Recognition. pp. 2321--2328 (2014)

\bibitem{khan2004cluster}
Khan, S.S., Ahmad, A.: Cluster center initialization algorithm for k-means
  clustering. Pattern recognition letters  \textbf{25}(11),  1293--1302 (2004)

\bibitem{le2018deepvq}
Le~Tan, D.K., Le, H., Hoang, T., Do, T.T., Cheung, N.M.: Deepvq: A deep network
  architecture for vector quantization. In: Proceedings of the IEEE Conference
  on Computer Vision and Pattern Recognition Workshops. pp. 2579--2582 (2018)

\bibitem{lecun1998gradient}
LeCun, Y., Bottou, L., Bengio, Y., Haffner, P.: Gradient-based learning applied
  to document recognition. Proceedings of the IEEE  \textbf{86}(11),
  2278--2324 (1998)

\bibitem{li2009constrained}
Li, Z., Liu, J.: Constrained clustering by spectral kernel learning. In:
  Computer vision, 2009 IEEE 12th international conference on. pp. 421--427.
  IEEE (2009)

\bibitem{liu2017balanced}
Liu, H., Han, J., Nie, F., Li, X.: Balanced clustering with least square
  regression. (2017)

\bibitem{macqueen1967some}
MacQueen, J., et~al.: Some methods for classification and analysis of
  multivariate observations. In: Proceedings of the fifth Berkeley symposium on
  mathematical statistics and probability. vol.~1, pp. 281--297. Oakland, CA,
  USA (1967)

\bibitem{mahajan2012planar}
Mahajan, M., Nimbhorkar, P., Varadarajan, K.: The planar k-means problem is
  {NP}-hard. Theoretical Computer Science  \textbf{442},  13--21 (2012)

\bibitem{malinen2014balanced}
Malinen, M.I., Fr{\"a}nti, P.: Balanced k-means for clustering. In: Joint IAPR
  International Workshops on Statistical Techniques in Pattern Recognition
  (SPR) and Structural and Syntactic Pattern Recognition (SSPR). pp. 32--41.
  Springer (2014)

\bibitem{pedregosa2011scikit}
Pedregosa, F., Varoquaux, G., Gramfort, A., Michel, V., Thirion, B., Grisel,
  O., Blondel, M., Prettenhofer, P., Weiss, R., Dubourg, V., et~al.:
  Scikit-learn: Machine learning in python. Journal of machine learning
  research  \textbf{12}(Oct),  2825--2830 (2011)

\bibitem{pena1999empirical}
Pena, J.M., Lozano, J.A., Larranaga, P.: An empirical comparison of four
  initialization methods for the k-means algorithm. Pattern recognition letters
   \textbf{20}(10),  1027--1040 (1999)

\bibitem{rifkin2007notes}
Rifkin, R.M., Lippert, R.A.: Notes on regularized least squares  (2007)

\bibitem{wagstaff2001constrained}
Wagstaff, K., Cardie, C., Rogers, S., Schroedl, S.: Constrained k-means
  clustering with background knowledge. In: Proceedings of the Eighteenth
  International Conference on Machine Learning. pp. 577--584. Citeseer (2001)

\bibitem{wright1999numerical}
Wright, S., Nocedal, J.: Numerical optimization. Springer Science
  \textbf{35}(67-68), ~7 (1999)

\bibitem{yang2016towards}
Yang, B., Fu, X., Sidiropoulos, N.D., Hong, M.: Towards k-means-friendly
  spaces: Simultaneous deep learning and clustering. arXiv preprint
  arXiv:1610.04794  (2016)

\bibitem{zhu2010data}
Zhu, S., Wang, D., Li, T.: Data clustering with size constraints.
  Knowledge-Based Systems  \textbf{23}(8),  883--889 (2010)

\end{thebibliography}
\end{document}